\documentclass[runningheads]{llncs}
\usepackage[T1]{fontenc}
\usepackage{graphicx}
\usepackage{amsmath}
\usepackage{amssymb}
\usepackage{array}      
\usepackage{stfloats}
\usepackage{cuted}      
\usepackage{caption}    
\usepackage{diagbox}
\usepackage{booktabs}
\usepackage{adjustbox}  
\usepackage{multirow}
\usepackage{subcaption}
\usepackage{placeins}

\newcommand{\commentout}[1]{}

\begin{document}
\title{LayerGS: Decomposition and Inpainting of Layered 3D Human Avatars via 2D Gaussian Splatting}
\titlerunning{LayerGS}
%
\author{Yinghan Xu,  John Dingliana}
\authorrunning{Y. Xu et al.}
%
\institute{Trinity College Dublin, Dublin, Ireland\\
\email{\{yixu,john.dingliana\}@tcd.ie}}
\maketitle              
\begin{abstract}
We propose a novel framework for decomposing arbitrarily posed humans into animatable multi-layered 3D human avatars, separating the body and garments. Conventional single-layer reconstruction methods lock clothing to one identity, while prior multi-layer approaches struggle with occluded regions. We overcome both limitations by encoding each layer as a set of 2D Gaussians for accurate geometry and photorealistic rendering, and inpainting hidden regions with a pretrained 2D diffusion model via score-distillation sampling (SDS). Our three-stage training strategy first reconstructs the coarse canonical garment via single-layer reconstruction, followed by multi-layer training to jointly recover the inner-layer body and outer-layer garment details. Experiments on two 3D human benchmark datasets (4D-Dress, Thuman2.0) show that our approach achieves better rendering quality and layer decomposition and recomposition than the previous state-of-the-art, enabling realistic virtual try-on under novel viewpoints and poses, and advancing practical creation of high-fidelity 3D human assets for immersive applications. Our code is available at \url{https://github.com/RockyXu66/LayerGS}.

\keywords{3D Human \and Virtual Try-on \and Gaussian Splatting.}
\end{abstract}
\section{Introduction}
\label{sec:intro}

Creating realistic digital representations of humans is essential in enhancing immersiveness of extended reality applications. One popular application is that of realistic 3D virtual try-on, which poses challenges due to limited availability of 3D garment assets and difficulty in capturing suitable 3D human models. Several methods have been developed to reconstruct high-quality human figures using single image~\cite{9010814}, monocular~\cite{Hu2023GaussianAvatarTR}, multi-view~\cite{Li2024AnimatableGL}, or RGBD video input~\cite{10.1007/978-3-031-19809-0_18}. However, most existing works focus on single-layer representations and do not separate garments from the inner body, limiting their application in virtual try-on. Due to lack of information in the capture input, occluded body parts are neglected, while garments are challenging to model accurately due to nonlinear material properties and dynamic deformations. 
To address this without delving into garment modeling, Kim et al. introduced GALA \cite{10656933}, a framework that utilizes the general knowledge from 2D diffusion models as a prior for geometry and appearance in the context of multi-layer human inpainting and generation. Given a static, single-layer 3D human scan, GALA can train and extract multi-layer models separating the body and garments. However, decomposition quality depends greatly on the original 3D scan. Although a 3D human model can be estimated from a single image~\cite{9010814}, a high-quality scan may require over one hundred cameras~\cite{Wang20244DDRESSA4}, limiting the potential usage of their method.

\begin{figure*}[t]
    \centering
    \includegraphics[width=1.0\textwidth]{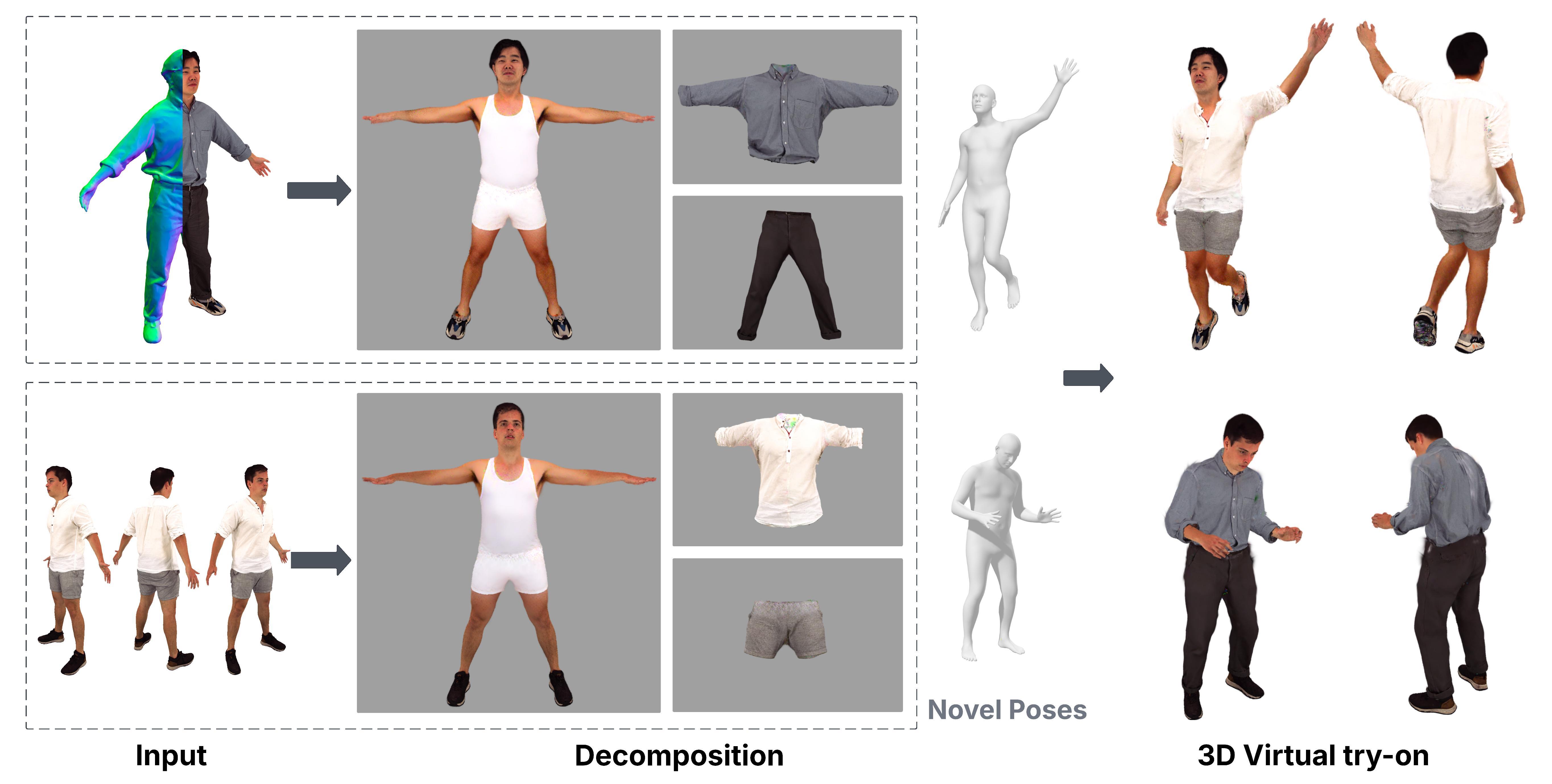}
    \caption{Given a 3D human scan or multi-view images of a static person, our framework decomposes and inpaints the subject into multiple canonical Gaussian layers for animation and 3D virtual try-on. \vspace{1em}}
\end{figure*}

In this paper, we present a framework for creating a multi-layer human representation from multi-view RGB images in static pose. Unlike previous methods, our approach is not limited to reconstructing a multi-layer human solely from a 3D scan, allowing realistic static human reconstruction from monocular video. Similar to GALA, we utilize a 2D diffusion model as a geometry and appearance prior, but %
in contrast to GALA's vertex color and DMTet \cite{shen2021dmtet} representation, we use the rendering and modeling advantages of 2D Gaussian Splatting (2DGS) \cite{Huang2DGS2024}. Experiments show that our method achieves improved image quality after decomposition and re-composition, enabling garment swapping between subjects and creating a realistic inner body and outer garment appearance for occluded parts of human individuals.

In summary, our key contributions are as follows:
\begin{itemize}
    \item We propose a framework that decomposes a realistic 3D human into multiple 3D layers, achieving better results than existing state-of-the-art methods.
   \item Instead of relying on 3D scans, we utilize 2D Gaussian Splatting to reconstruct and inpaint the 3D human model only from multi-view RGB images.
    \item Additionally, we explicitly model these decomposed layers with 3D meshes, making them suitable for 3D virtual try-on applications.       
\end{itemize}

\section{Related Work}
\label{sec:relatedwork}

\subsection{Single-layer Human Modeling}

PIFu \cite{9010814} reconstructs humans by training a 3D implicit field to capture geometry information from a single 2D image. A single-layer geometry is extracted by marching cubes, and vertex colors are predicted. SIFU~\cite{Zhang_2024_CVPR} introduces 3D consistent texture refinement for high-quality 3D clothed human models. HumanSplat~\cite{pan2024humansplat} generates a 3D Gaussian human from a single image with a 2D multi-view diffusion model and human structure prior. 
IP-Net~\cite{bhatnagar2020ipnet} uses point clouds as input to generate a single-layer registered animatable avatar with clothes. Some methods \cite{Chen2021SNARFDF} \cite{Ma2021ThePO} \cite{Shen2023XAvatarEH} use a sequence of 3D scans of the human as input to generate single-layer pose-dependent clothing deformation. \cite{Ma2021ThePO} focus on single-layer geometry reconstruction and representation, while X-Avatar~\cite{Shen2023XAvatarEH} extend this to geometry with textures. However, all these methods treat the human avatar as a single layer, ignoring the geometry and texture information under the clothing. These approaches also limit the ability for virtual try-on applications and clothes swapping between different subjects.

\subsection{Multi-layer Human Modeling}

Modeling humans with disentangled clothing layers is critical for editable avatars. SMPLicit~\cite{Corona2021SMPLicitTG} fits human cloth on top of SMPL~\cite{Loper2023SMPLAS} parameter models with a given 2D image but lacks texture generation. CaPhy~\cite{Su2023CaPhyCP} generates the human and garment template from a sequence of 3D scans while learning the physical properties of animatable garments. LayGA~\cite{Lin2024LayGALG} use multi-view videos as input and model the body and garment separately. The garment properties are learned from the temporal observations. However, acquiring such dense video data is costly, and the method fail to accurately reconstruct the occluded inner body, leading to artifacts when the avatar wears shorter clothes than the original. DeClothH~\cite{nam2025decloth} reconstructs a multi-layer 3D clothed human from a single image using a specifically designed diffusion model but struggles to maintain high-fidelity 3D consistency. LayerAvatar~\cite{zhang2025disentangled} introduces a feed-forward diffusion model that generates disentangled clothed avatars using a layered UV feature plane representation. Despite its speed, LayerAvatar focuses on generating new identities from latent noise rather than editing existing avatars under specific user guidance.

\subsection{Virtual Try-on}

Several studies address 2D image inpainting by utilizing the capabilities of large-scale pre-trained text-to-image models, such as Stable Diffusion~\cite{Rombach2021HighResolutionIS}. ControlNet-Inpainting~\cite{10377881} can generate realistic 2D human images based on description prompts or 2D human poses. To specifically enhance garment-related tasks, \cite{Xie2023GPVTONTG} \cite{Zhu2024MMVM} focus on virtual try-on and editing of garments in 2D. However, all remain limited to fixed viewpoints and lack immersive 3D support.

Researchers lift 2D inpainting into 3D by using a 2D diffusion model as a prior and combining it with 3D representations. For example, Instruct-NRF2NeRF~\cite{Haque2023InstructNeRF2NeRFE3} enables instruction-based editing of Neural Radiance Fields~\cite{mildenhall2020nerf}. GaussianEditor~\cite{Chen2023GaussianEditorSA} provides 3D editing capabilities for 3DGS assets, leveraging 2D diffusion models. VTON360~\cite{He2025VTON3H} reformulates 2D VTON as a multi-view 2D inpainting task. It projects the 3D human into 2D views, applies a diffusion-based inpainter conditioned on garment images, and then reconstructs the edited views back into 3D using Gaussian Splatting. While effective for specific views, this editing-based pipeline lacks explicit layer decomposition, often leading to inconsistency.

\section{Method}
\begin{figure*}[t]
    \centering
    \includegraphics[width=\textwidth]{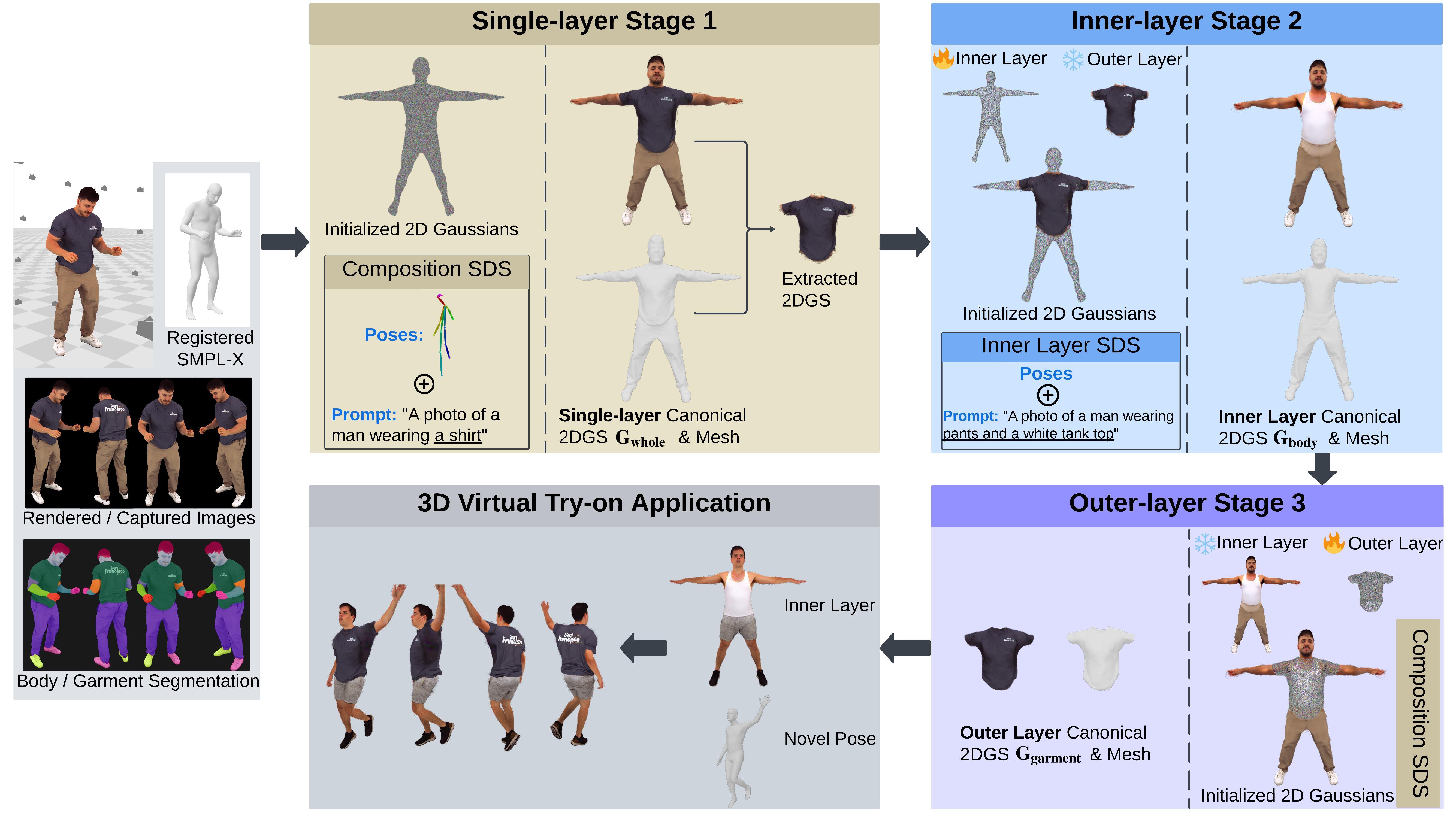}
    \caption{
    \textbf{Overview.} (Left) input multi-view RGB and segmentation masks. In \textbf{Stage 1}, we learn a single-layer canonical set of 2D Gaussians by integrating with a composition SDS loss and extract a coarse garment mesh by segmentation. In \textbf{Stage 2}, we decompose and optimize an inner Gaussian layer (body and inner garments) while keeping the outer layer fixed. In \textbf{Stage 3}, we refine the outer garment layer with the inner layer frozen. Both resulting Gaussian layers are attached to the mesh to enable high-fidelity 3D virtual try-on under novel poses.}
    \label{fig:overview}
\end{figure*}
We reconstruct Gaussian human avatar layers $G_{body}$ and $G_{garment}$ from captured 2D images of a static human, enabling realistic rendering results from novel views and poses (Fig.~\ref{fig:overview}), using a three-stage pipeline. In Stage 1, we train a single-layer canonical human avatar for the coarse outer layer garment reconstruction and inpainting. In Stage 2, we freeze the outer layer garment from Stage 1 and optimize the canonical inner layer. In Stage 3, we freeze the canonical inner layer and train the refined outer layer to address the rendering artifacts caused by inter-layer overlaps. Finally, we extract a 3D mesh from each layer and attach the corresponding Gaussians to this mesh, enabling seamless 3D virtual try-on by recomposition of 3D assets under novel poses. Stages~1--3 are detailed in Sec~\ref{subsec:stage1}, Sec~\ref{subsec:stage2},~and Sec~\ref{subsec:stage3}.

\subsection{Single-layer Canonical Avatar Reconstruction and Inpainting}\label{subsec:stage1}
Unlike GALA~\cite{10656933}, which requires a static human associated with a 3D mesh, our approach only needs 2D multi-view images, which can be captured through a monocular video with a camera moving spherically around a human subject in a random posture, or rendered from spherical virtual camera viewpoints placed around a 3D human scan. 
The SMPL(-X) parameters are given or fitted. We sample 30k points from the canonical SMPL(-X) mesh surface and initialize 2D Gaussians. A 3D grid linear skinning weight (LBS) is pre-computed from SMPL(-X) parameters and interpolated to compute skinning weights for 2D Gaussians.

In this stage, the posed 2D Gaussian representation of a human is rendered as a single layer. The attributes of the 2D Gaussians \(\mathbf{G_{whole}}\) can be optimized using 2D ground-truth images. The RGB image loss function $\mathcal{L}^{rgb}$ is as follows:
\begin{equation}
    \mathcal{L}_{whole}^{rgb} = (1-\lambda_{c}) \mathcal{L}_{whole}^{L1} + \lambda_{c} \mathcal{L}_{whole}^{\text{D-SSIM}}
\end{equation}
where the losses $\mathcal{L}_{whole}^{L1}$ and $\mathcal{L}_{whole}^{\text{D-SSIM}}$ represents $L1$ loss and D-SSIM loss, respectively, for RGB images $g((\mathbf{G_{whole}}), \mathbf{P})$ rendered by the single layer, with the weight coefficient $\lambda_{c}$. $\mathbf{P}$ denotes the camera model.

We apply two regularizations following the original 2DGS work to obtain more accurate geometries. Depth distortion loss \(\mathcal{L}_d\) is applied to mitigate the spread of Gaussians along the intersections. Normal consistency loss \(\mathcal{L}_n\) is applied to ensure intersected 2D splats share the same normal, which is aligned with the actual surface. Depth distortion loss is as follows:
\begin{equation}
    \mathcal{L}_{whole}^d = \sum_{i,j} \omega_i \omega_j |z_i - z_j|
\end{equation}
where $\omega_i$ is the blending weight of the $i$-th intersection and $z_i$ is the depth of the intersection points. Normal consistency loss is as follows:
\begin{equation}
    \mathcal{L}_{whole}^n = \sum_{i} \omega_i (1 - \text{n}_i^T \text{N})
\end{equation}
where $\omega_i$ is the blending weights of the $i$-th intersection, $\text{n}_i$ represents the normal of the 2D splat, and $\text{N}$ is the normal estimated by the gradient of the depth map.

For a fair comparison with GALA, we use the same diffusion model, Stable Diffusion V1.5~\cite{Rombach2021HighResolutionIS}, for score distillation sampling (SDS) loss~\cite{poole2023dreamfusion}. The diffusion model is augmented with ControlNet~\cite{10377881} and OpenPose~\cite{8765346}. The SDS loss is applied to the rendered image, enabling inpainting of occluded areas in the single layer \(\mathbf{G_{whole}}\). The SDS text prompt is specified manually for each subject (e.g., ""a photo of a man wearing a t-shirt""). Gaussians in occluded areas are optimized through densification and pruning, particularly in areas between the upper arm and the torso. The gradient of SDS guidance loss is as follows:
\begin{equation}
\resizebox{.9\hsize}{!}{$
    \nabla_{\mathbf{G_{\text{whole}}}} \mathcal{L}_{\text{whole}}^{\text{SDS}} = \mathbb{E}_{\epsilon, t} 
\left[ w_t \left( \epsilon_{\phi} (\mathbf{z}_t^{\text{whole}}; y_{\text{whole}}, t) - \epsilon \right) 
\frac{\partial \mathbf{x_{whole}}}{\partial \mathbf{G_{whole}}} \right]
$}
\end{equation}
where $\mathbf{x_{whole}}$ represents the image rendered using the single layer parameters $\mathbf{G_{whole}}$. The variable $\mathbf{z}_t^{\text{whole}}$ refers to the corresponding noisy image. The text embedding $y_{\text{whole}}$ is derived from the text prompt describing the subject wearing the garment. The total loss in Stage 1 is the weighted sum of the four losses.

After obtaining the whole single-layer canonical avatar, we use the same technique in 2DGS to extract the mesh. Specifically, depth maps of the training views are rendered and fused to a mesh by utilizing truncated signed distance fusion (TSDF). Then, we render multiple-view 2D images for this canonical mesh, following a 2D segmentation and 3D votes to get the segmented canonical garment mesh. Assuming 2D Gaussians are spread on the mesh surface, we get the coarse canonical outer layer garment by setting a closest distance threshold $\alpha=0.015$ from the center position of Gaussians to the mesh.

\subsection{Inner Layer Canonical Avatar Reconstruction and Inpainting}\label{subsec:stage2}
We initialize the canonical inner layer by sampling points from the SMPL(-X) mesh surface. During training, we freeze the outer layer garment from the previous step. Apart from RGB image loss, normal consistency loss, and score distillation sampling (SDS) loss for the inner layer 2D Gaussians, we also apply a 2D segmentation loss for the body, garment, and background. The 2D segmentation labels $M \in \{0,1\}^{H\times W}$ are employed for each layer. We divide depth distortion loss into seen and occluded areas for the inner layer based on 2D segmentation labels. The seen area is computed as normal while we create a dummy SMPL(-X) 2D Gaussian set and render it together with the optimizable inner layer to guide the depth information for the occluded area. These segmentation and depth regularizations are essential for guiding the shape of the inner layer body from 2D observations, restricting the body shape from going beyond the garment, and fixing inaccurate intersections between the garment and body. The loss functions are as follows:
\begin{equation}
    \mathcal{L}_\text{body}^{rgb} = (1-\lambda_{c}) \mathcal{L}_{\text{body}}^{L1} + \lambda_{c} \mathcal{L}_{\text{body}}^{\text{D-SSIM}}
\end{equation}
\begin{equation}
\resizebox{.7\hsize}{!}{$
\mathcal{L}_{\mathrm{seg}}
= \sum_{k}\bigl\|M_{k}\odot(\mathbf{S}-\mathbf{S}_{k}^{gt})\bigr\|_{1}
\quad\text{where }k\in\{\mathrm{body},\,\mathrm{garment},\,\mathrm{bg}\}
$}
\end{equation}
\begin{equation}
    \mathcal{L}_{body}^d = \sum_{i_s,j} \omega_{i_s} \omega_j |z_{i_s} - z_j| + \sum_{i_o,j} \omega_{i_o} \omega_j |z_{i_o} - z_j|
\end{equation}
\begin{equation}
\resizebox{.8\hsize}{!}{$
    \nabla_{\mathbf{G_{\text{body}}}} \mathcal{L}_{\text{body}}^{\text{SDS}} = \mathbb{E}_{\epsilon, t} 
\left[ w_t \left( \epsilon_{\phi} (\mathbf{z}_t^{\text{body}}; y_{\text{body}}, t) - \epsilon \right) 
\frac{\partial \mathbf{x_{body}}}{\partial \mathbf{G_{body}}} \right]
$}
\end{equation}
where the losses $\mathcal{L}_{body}^{l1}$ and $\mathcal{L}_{body}^{\text{D-SSIM}}$ represents $l1$ loss and D-SSIM loss respectively for masked RGB images $g((\mathbf{G_{body}, \mathbf{G_{garment}}}), \mathbf{P})$ rendered by integrating the body and garment layers. The target segmentation image $\mathbf{S}^{gt}$ is assigned a uniform color, represented as $\mathbf{S}^{gt}=v\cdot \mathbf{1}^{H \times W}$, where $v$ is an arbitrary RGB value for different layers. Aside from RGB images, we render segmentation images $\mathbf{S} = g((\mathbf{G'_{body}}, \mathbf{G'_{garment}}), \mathbf{P})$, where $\mathbf{G'} = (\mu, c', s, \alpha, q)$. Here, the color for each 2D Gaussian is set to match the corresponding layer color, $c'_{body} = v_{body}$, $c'_{garment} = v_{garment}$. For depth distortion loss $\mathcal{L}_{body}^d$, we define masks for the $i$-th intersection: $i_s$ denotes the seen region of the body, while $i_o$ denotes the occluded region, obtained by masking with the garment and computed using the dummy body shape. The text prompt for SDS loss for the inner layer is given according to the user requirement (e.g., "a photo of a man wearing pants and a white tank top"). Similar to Stage 1, the Stage 2 total loss is computed as a weighted sum of the RGB reconstruction, normal consistency, depth distortion, and SDS guidance losses, with additional 2D segmentation losses for the inner layer, outer layer, and the background.

\begin{table*}[t]
    
    \begin{center}

    \begin{adjustbox}{max width=\textwidth}


        \begin{tabular}{l|ccc|ccc|ccc}
        \toprule
        \multirow{2}{*}[-0.75ex]{Method}
        & \multicolumn{3}{c|}{00122-Inner-Take8} 
        & \multicolumn{3}{c|}{00127-Inner-Take5}
        & \multicolumn{3}{c}{00152-Inner-Take4} \\
        \cmidrule(lr){2-4}\cmidrule(lr){5-7}\cmidrule(lr){8-10}
        & SSIM$\uparrow$ & PSNR$\uparrow$ & LPIPS$\downarrow$
        & SSIM$\uparrow$ & PSNR$\uparrow$ & LPIPS$\downarrow$
        & SSIM$\uparrow$ & PSNR$\uparrow$ & LPIPS$\downarrow$ \\
        \midrule
        GALA w/ 3D scan     
            & 0.9801 & 34.60 & 0.0257 
            & 0.9766 & 32.53 & 0.0321
            & 0.9805 & 33.38 & 0.0268 \\
        GALA w/o 3D scan     
            & 0.9697 & 31.70 & 0.0382 
            & 0.9684 & 31.44 & 0.0405
            & 0.9715 & 31.74 & 0.0381 \\
        \midrule
        Ours w/o 3D scan              
            & \textbf{0.9873} & \textbf{35.40} & \textbf{0.0195}
            & \textbf{0.9880} & \textbf{35.93} & \textbf{0.0191}
            & \textbf{0.9869} & \textbf{35.91} & \textbf{0.0214} \\
        \bottomrule
        \end{tabular}
    \end{adjustbox}

    \begin{adjustbox}{max width=\textwidth}
    
        \begin{tabular}{l|ccc|ccc|ccc}
        \toprule
        \multirow{2}{*}[-0.75ex]{Method}
        & \multicolumn{3}{c|}{00174-Inner-Take10} 
        & \multicolumn{3}{c|}{00175-Inner-Take4}
        & \multicolumn{3}{c}{00190-Inner-Take2} \\
        \cmidrule(lr){2-4}\cmidrule(lr){5-7}\cmidrule(lr){8-10}
        & SSIM$\uparrow$ & PSNR$\uparrow$ & LPIPS$\downarrow$
        & SSIM$\uparrow$ & PSNR$\uparrow$ & LPIPS$\downarrow$
        & SSIM$\uparrow$ & PSNR$\uparrow$ & LPIPS$\downarrow$ \\
        \midrule
        GALA w/ 3D scan
            & 0.9801 & 34.93 & 0.0340 
            & 0.9792 & 31.18 & 0.0292 
            & 0.9732 & 33.37 & 0.0250 \\
        GALA w/o 3D scan     
            & 0.9718 & 30.91 & 0.0453
            & 0.9674 & 28.88 & 0.0448 
            & 0.9628 & 31.25 & 0.0346 \\
        \midrule
        Ours w/o 3D scan        
            & \textbf{0.9866} & \textbf{36.09} & \textbf{0.0306}
            & \textbf{0.9882} & \textbf{36.67} & \textbf{0.0207} 
            & \textbf{0.9857} & \textbf{35.02} & \textbf{0.0153} \\
        \bottomrule
        \end{tabular}
    \end{adjustbox}

    \end{center}
    \caption{\small \textbf{Quantitative recomposition evaluation on the 4D-Dress dataset.}}
    \label{tab:4d-dress-recomposition-evaluation}
\end{table*}

\subsection{Outer Layer Refinement}\label{subsec:stage3}
Since the outer layer is reconstructed and inpainted as a single layer in Stage 1, some artifacts may occur when combining the inner layer with the outer layer, especially in occluded area optimized using SDS. To mitigate this issue, we freeze 2D Gaussians of the inner layer and only refine the outer layer. Initialization is peformed by sampling points from the coarse canonical outer layer generated from Sec~\ref{subsec:stage1}. To further accelerate the training process, we employ a dummy-guidance strategy similar to Sec~\ref{subsec:stage2}, using the coarse outer layer mesh as the dummy layer. To reduce rendering artifacts caused by severe scaling deformation during virtual try-on, we apply a scaling constraint in densification: 2D Gaussians are split at a manually set scaling threshold of 0.01, and scaling values are clipped to 0.01 after splitting.

The training losses include masked RGB image loss for the outer layer area, normal consistency loss, depth distortion loss, 2D segmentation losses, and composition SDS loss with the same text prompt as in Stage 1. The normal consistency loss and depth distortion loss are applied to both (a) the composition of the inner and outer layers and (b) the composition of the dummy and outer layers. The Stage 3 total loss is the weighted sum of all these losses.

\subsection{Mesh Extraction and Virtual Try-on}
We employ a mesh-driven approach for 3D virtual try-on applications. First, we extract 3D meshes for both the canonical body layer and the canonical garment layer from corresponding 2D Gaussians assets, respectively. Next, we attach 2D Gaussians to their nearest triangles on the canonical mesh. To accommodate novel poses, we apply linear blend skinning (LBS) for the canonical mesh transformation and Laplacian deformation to the outer layer to resolve potential penetration issues between inner and the outer layers. Finally, 2D Gaussians are transformed from the canonical mesh to the target mesh.

\section{Experiments}

\subsection{Preprocessing}\label{subsec:preprocessing}
To our knowledge, GALA is the only existing method that decomposes realistic 3D humans into multiple layers with inpainting.
For a fair comparison, we adopt a preprocessing pipeline analogous to GALA. To enhance adaptability to arbitrary multi-view image sets, we further estimate camera intrinsics and extrinsics jointly with human shape and pose parameters from scratch. Given multi-view RGB images of a static human, we calculate the camera extrinsics and intrinsics using COLMAP~\cite{schoenberger2016sfm}. Next, we utilize an off-the-shelf 2D segmentation model~\cite{Khirodkar2024SapiensFF} to generate annotation masks for both body and garment components. The body shape, represented as $\beta \in \mathbb{R}^{10}$, and the body poses, denoted as $\theta \in \mathbb{R}^{144}$, for the subject are obtained by optimizing the SMPL(-X) parameters based on the multi-view images and the corresponding camera poses.
\begin{table*}[t]
    \begin{center}
    
    \begin{adjustbox}{max width=0.8\linewidth}
    
        \begin{tabular}{l|cc|cc|cc}
        \toprule
        \multirow{2}{*}[-0.75ex]{Method}
        & \multicolumn{2}{c|}{00122-Inner-Take8} 
        & \multicolumn{2}{c|}{00127-Inner-Take5}
        & \multicolumn{2}{c}{00152-Inner-Take4} \\
        \cmidrule(lr){2-3}\cmidrule(lr){4-5}\cmidrule(lr){6-7}
        & CLIP$\uparrow$ & IR$\uparrow$ 
        & CLIP$\uparrow$ & IR$\uparrow$ 
        & CLIP$\uparrow$ & IR$\uparrow$ \\
        \midrule
        GALA w/ 3D scan     
            & \textbf{30.50} & 0.313
            & 30.10 & 0.151
            & 30.11 & -0.513 \\
        GALA w/o 3D scan     
            & 30.08 & 0.175
            & 29.34 & 0.138
            & 29.90 & -0.638 \\
        \midrule
        Ours w/o 3D scan              
            & 30.48 & \textbf{0.534}
            & \textbf{30.97} & \textbf{0.217}
            & \textbf{31.08} & \textbf{-0.362} \\
        \bottomrule
        \end{tabular}
    
    \end{adjustbox}

    \begin{adjustbox}{max width=0.8\linewidth}
        \begin{tabular}{l|cc|cc|cc}
        \toprule
        \multirow{2}{*}[-0.75ex]{Method}
        & \multicolumn{2}{c|}{00174-Inner-Take10} 
        & \multicolumn{2}{c|}{00175-Inner-Take4} 
        & \multicolumn{2}{c}{00190-Inner-Take2} \\
        \cmidrule(lr){2-3}\cmidrule(lr){4-5}\cmidrule(lr){6-7}
        & CLIP$\uparrow$ & IR$\uparrow$ 
        & CLIP$\uparrow$ & IR$\uparrow$ 
        & CLIP$\uparrow$ & IR$\uparrow$ \\
        \midrule
        GALA w/ 3D scan 
            & 31.27 & 1.715
            & 32.21 & 1.783 
            & 31.30 & 0.689 \\
        GALA w/o 3D scan 
            & 30.30 & 1.641
            & 31.18 & 1.676
            & 29.64 & 0.491 \\
        \midrule
        Ours w/o 3D scan
            & \textbf{31.82} & \textbf{1.767}
            & \textbf{33.26} & \textbf{1.843} 
            & \textbf{33.34} & \textbf{0.993} \\
        \bottomrule
        \end{tabular}

    \end{adjustbox}

    \end{center}
    \caption{\small \textbf{Quantitative inpainting evaluation on the 4D-Dress dataset.}}
    \label{tab:4d-dress-inpainting-evaluation}
\end{table*}
\begin{figure}[t]
    \centering
    \setlength{\extrarowheight}{0pt}
    \renewcommand{\arraystretch}{1}
    
    \begin{tabular}{@{} m{0.9cm} @{} m{0.92\linewidth} @{}}
        {\centering\scriptsize \textbf{GALA \newline w/ scan}} & \includegraphics[width=0.95\linewidth,keepaspectratio]{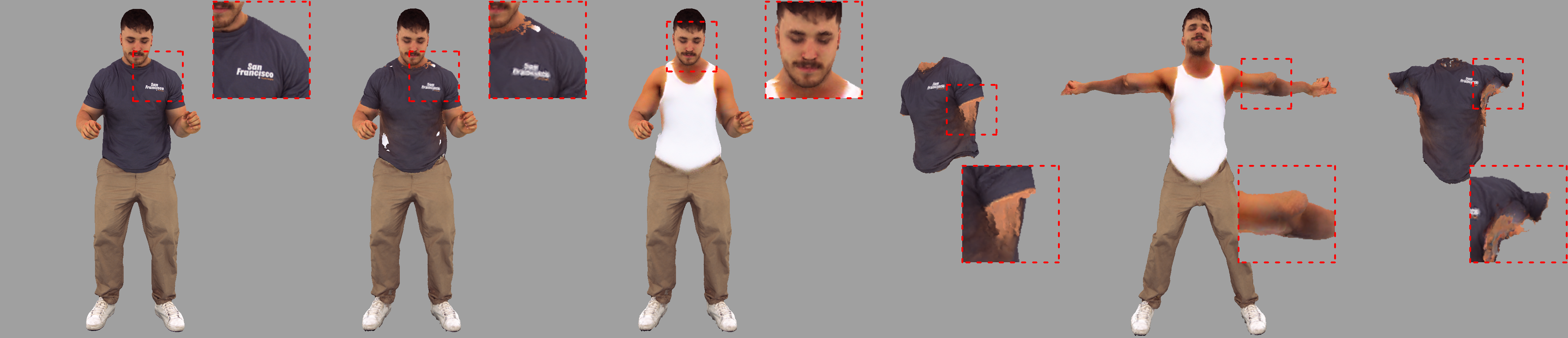} \\[0pt]
        {\centering\scriptsize \textbf{GALA \newline w/o scan}} & \includegraphics[width=0.95\linewidth,keepaspectratio]{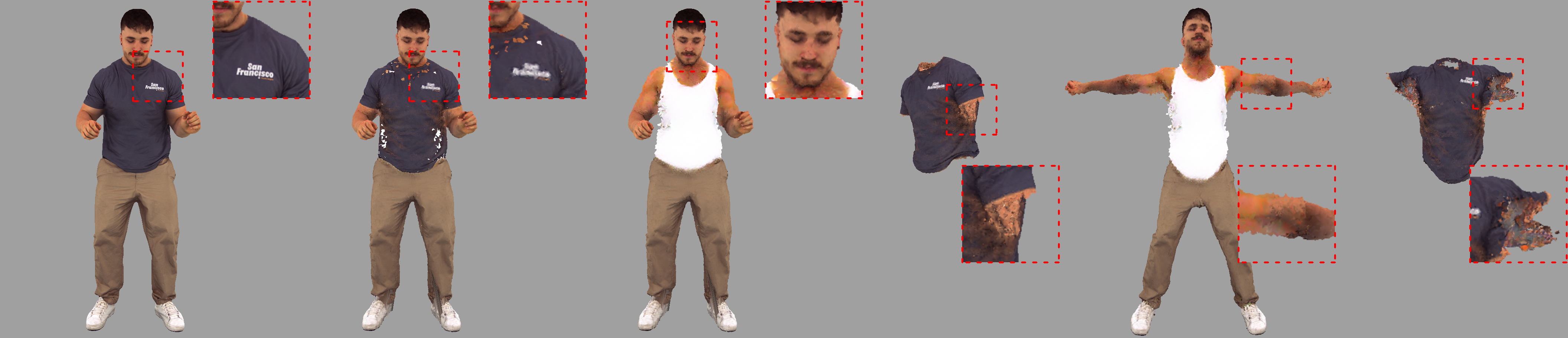} \\[0pt]
        {\centering\scriptsize \textbf{Ours \newline w/o scan}} & \includegraphics[width=0.95\linewidth,keepaspectratio]{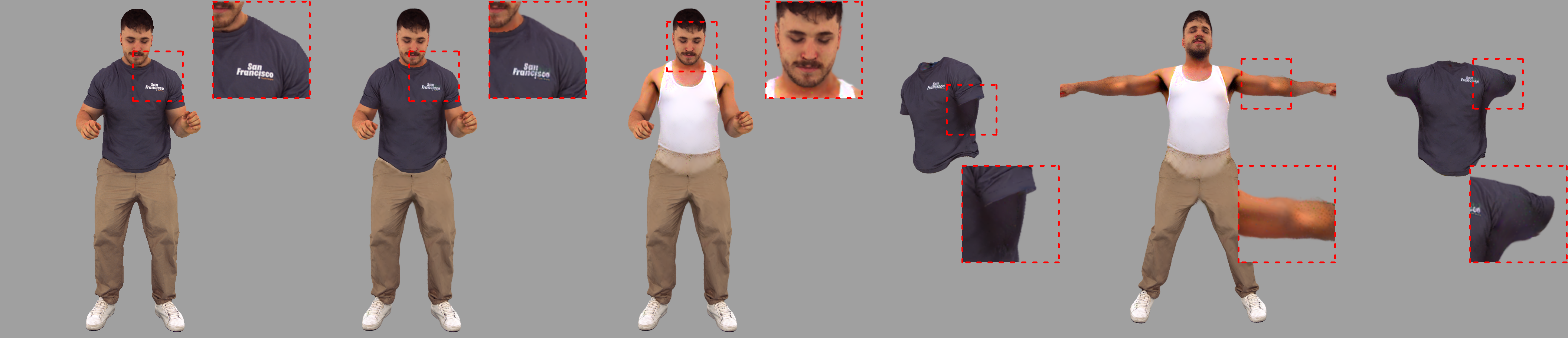} \\
    \end{tabular}
    \captionsetup{font=footnotesize}
    \caption{\textbf{Qualitative comparison between GALA (with and without 3D scan) and our method (without 3D scan).} The top row shows results from GALA trained with a 3D scan, the middle row from GALA trained without a 3D scan, and the bottom row from our approach, which also does not require a 3D scan. From left to right, each column shows: (1) the ground-truth image, (2) recomposed body and garment, (3) posed body, (4) posed garment, (5) canonical body, and (6) canonical garment.}
    \label{fig:4d-dress-qualitative-comparison}
\end{figure}
\subsection{Datasets and Metrics}
4D-Dress \cite{Wang20244DDRESSA4} is a real-world dataset of clothed humans that captures 64 human outfits with motion sequences. Each frame features a high-quality 3D textured scan with vertex-level semantic labels, along with corresponding garment meshes and fitted SMPL(-X) body meshes. Thuman2.0 Dataset~\cite{Yu2021Function4DRH} contains 526 reconstructed clothed human scans. In evaluation, there are two types of training data. 
The first consists of 72 images captured from uniformly distributed virtual camera positions around the subject. 
The second includes a ground-truth 3D scan as used in GALA's original pipeline, along with training data including normal images and segmentation masks rendered from random rotations and distances, as well as head and hand zoom-in views.
Our method and VTON360 use only the first type of data, while GALA's framework uses both. For evaluation, we render 72 test views per subject: for GALA, camera positions are uniformly distributed over the viewing sphere, while for VTON360 they are evenly spaced along a horizontal circle around the subject.

\noindent\textbf{Evaluation Metrics:} To compare with GALA, we use SSIM \cite{Wang2004ImageQA}, PSNR, and LPIPS \cite{Zhang2018TheUE} to evaluate reconstruction quality. CLIP~\cite{Radford2021LearningTV} and ImageReward (IR)~\cite{Xu2023ImageRewardLA} scores are used to evaluate the inpainting quality. Similar to VTON360, we calculate the average DINO Similarity~\cite{oquab2023dinov2} score between the reference image and the rendered test-view images of the 3D virtual try-on.

\noindent\textbf{Experimental Settings:} To compare with GALA, we select six subjects with diverse body shapes from the 4D-Dress dataset, each with a random pose sampled from its motion sequence. The garments in our study include t-shirts, long-sleeve shirts, and sweaters. To compare with VTON360, we select six subjects from Thuman2.0 to serve as the basis for inner-body reconstruction. In this setup, our method combines the six reconstructed garments from the 4D-Dress dataset with these inner bodies. In contrast, VTON360 performs virtual try-on using the same Thuman2.0 subjects but takes only the front and back garment images  from the corresponding 4D-Dress outfits as input. In all quantitative evaluations, we focus solely on upper-body garments for comparison purposes.

\subsection{Comparisons with GALA}
\noindent\textbf{Recomposition Results:}
Quantitative evaluations (Table~\ref{tab:4d-dress-recomposition-evaluation}) on the 4D-Dress dataset demonstrate that our method consistently surpasses GALA across multiple metrics, including SSIM, PSNR, and LPIPS. Specifically, our method can effectively reconstruct and inpaint the body and garment when the subject is wearing a tight garment or clothing with complex textures and patterns. In contrast, GALA relies heavily on the geometry of 3D scans and the fitted SMPL(-X) model. If the SMPL(-X) template is inaccurate, it can lead to flawed geometry reconstruction in GALA's first stage, resulting in overlapping rendering during composition. GALA addresses this with an additional refinement step, but penetration artifacts remain. As the refinement code is not available, we report comparisons using their results before this step. Qualitative examples illustrating these differences are shown in Fig.~\ref{fig:4d-dress-qualitative-comparison}.

\noindent\textbf{Inpainting Results:} 
We render the canonical inpainted inner body under test views and average the scores with text prompt "a man/woman wearing pants and a white tank top with black background". Quantitative results in Fig.~\ref{tab:4d-dress-inpainting-evaluation} show that our method has higher CLIP and ImageReward scores compared to GALA. As shown in Fig.~\ref{fig:4d-dress-qualitative-comparison}, our approach generates more photorealistic and coherent inpainting results, due to the combination of 2DGS and the diffusion model.

\subsection{Comparison with VTON360}
\begin{figure}[t]
    \centering
    \includegraphics[width=\linewidth]{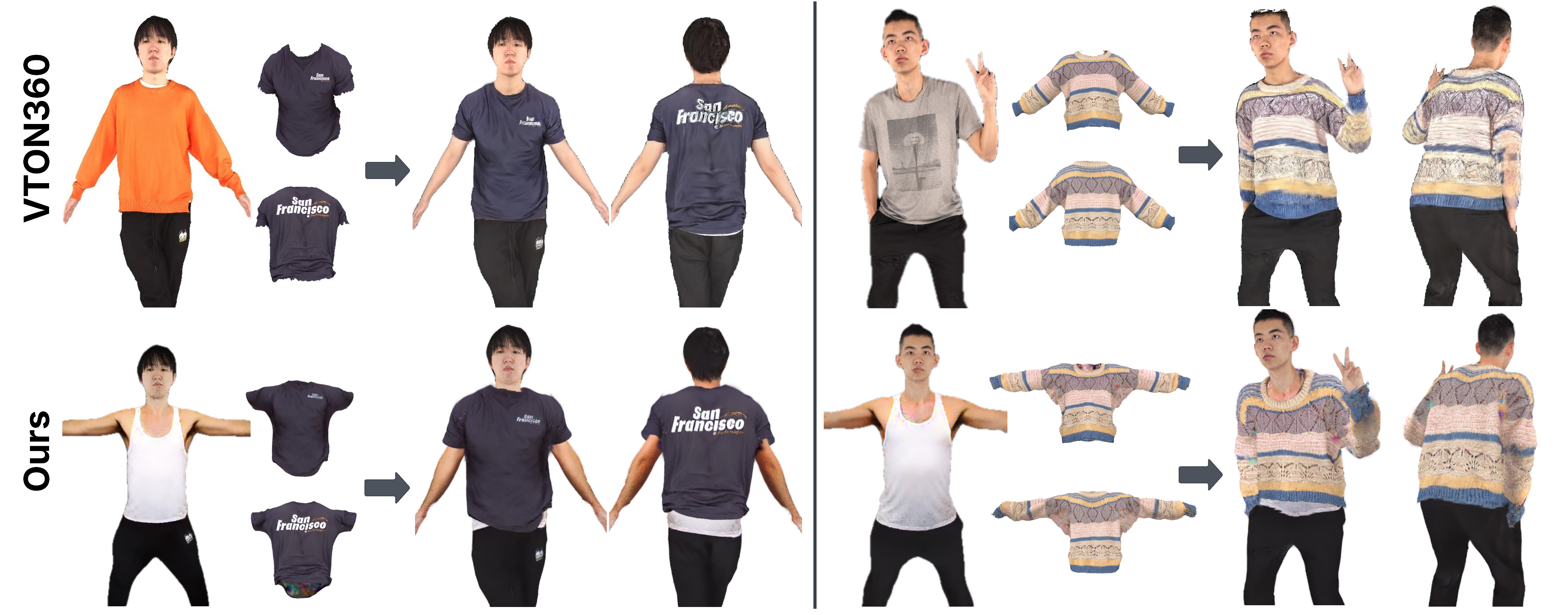}
    \captionsetup{font=footnotesize}
    \caption{\textbf{Qualitative comparison between VTON360 and our method.} Top: original subject, front/back garment images, and edited 3DGS renderings from VTON360. Bottom: canonical body, canonical garment, and recomposed results from our method.}
    \label{fig:vton360-comparison}
\end{figure}

In the evaluation, VTON360~\cite{He2025VTON3H} needs to train 6 subjects with 6 upper-body garments separately, while our method reconstructs all 12 subjects once and then recomposes the inner and outer layers arbitrarily. The quantitative results for the average DINO similarity score are shown in Table~\ref{tab:quantitative-comparison-vton360}. The qualitative comparisons in Fig.~\ref{fig:vton360-comparison} show that our method preserves high-frequency details, such as text on a t-shirt and patterns on a sweater, whereas VTON360 produces blurred results. Unlike our method which decomposes layers once for universal use, VTON360 requires re-training or optimizing the reconstruction for each specific garment-person pair.
\begin{table}[t]
    \centering
      \begin{tabular}{l|c}
      \toprule
      Method & DINO$_{sim}$ $\uparrow$  \\
      \midrule
      VTON360 & 0.455  \\
      Ours & \textbf{0.506} \\
      \bottomrule
      \end{tabular}
    \captionsetup{font=footnotesize}
    \caption{\small \textbf{Quantitative virtual try-on evaluation with VTON360 on the Thuman2.0/4D-Dress setup.}}
    \label{tab:quantitative-comparison-vton360}
\end{table}

\subsection{Custom Monocular Video Results} 
\begin{figure}[t]
    \centering
    \includegraphics[width=0.6\linewidth]{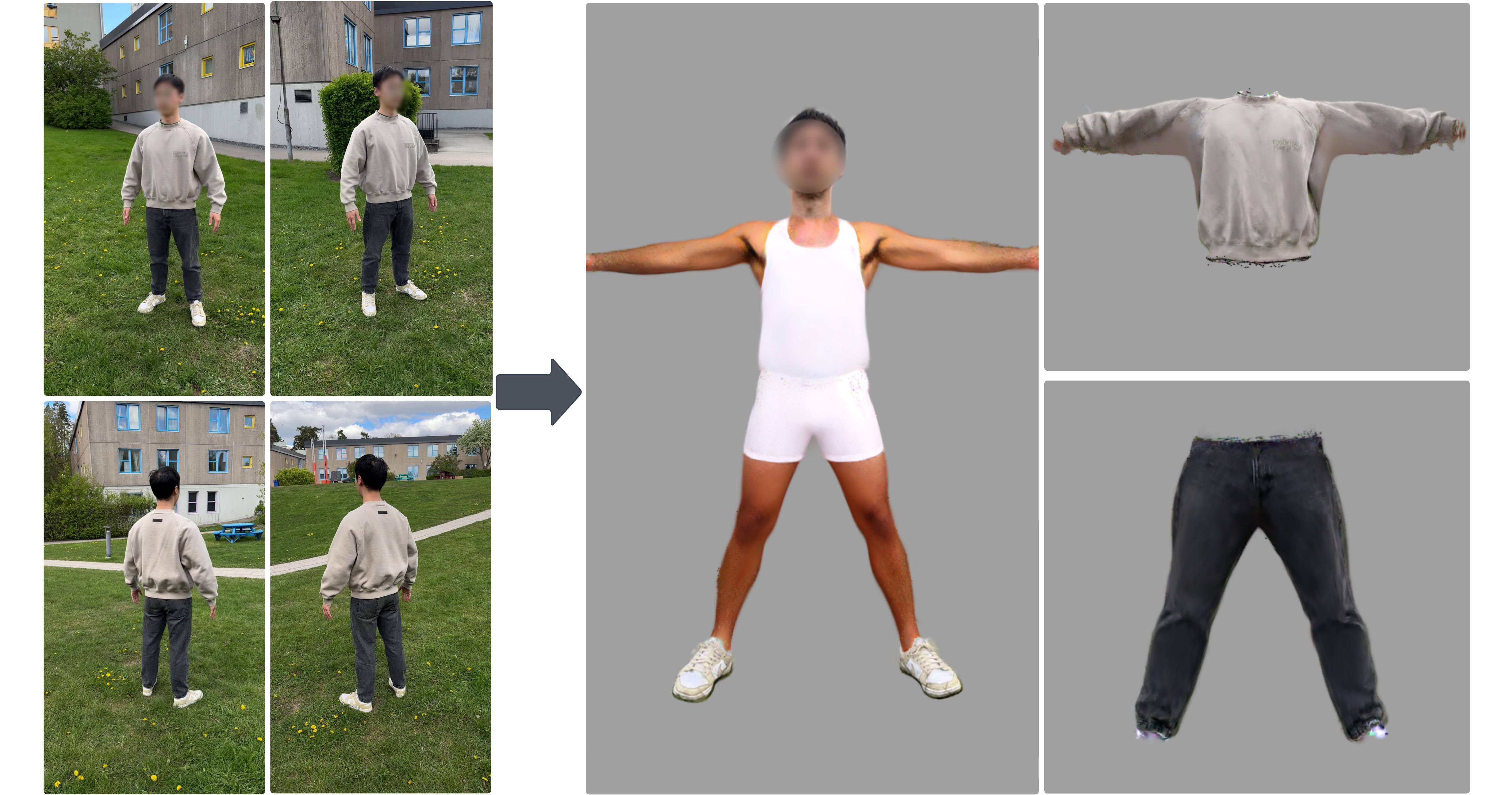}
    \captionsetup{font=footnotesize}
    \caption{\textbf{Custom monocular video demo for decompositions.}}
    \label{fig:monovideo-demo}
\end{figure}

An advantage of our method over GALA is its capability to decompose humans into multiple layers using only multi-view images without the requirement for explicit 3D information. To showcase this capability, we apply our system to custom data collected using a handheld monocular camera moved spherically around a static human subject in the scene. We extracted approximately 100 images from the video, and applied the preprocessing (Sec~\ref{subsec:preprocessing}) pipeline to these images. Our approach successfully reconstructs and inpaints layered 3D human avatars, as demonstrated in Fig.~\ref{fig:monovideo-demo}.

\subsection{Ablation Study}
\begin{figure*}[t]
    \centering
    \begin{subfigure}[b]{0.48\linewidth}
        \centering
        \includegraphics[width=\linewidth]{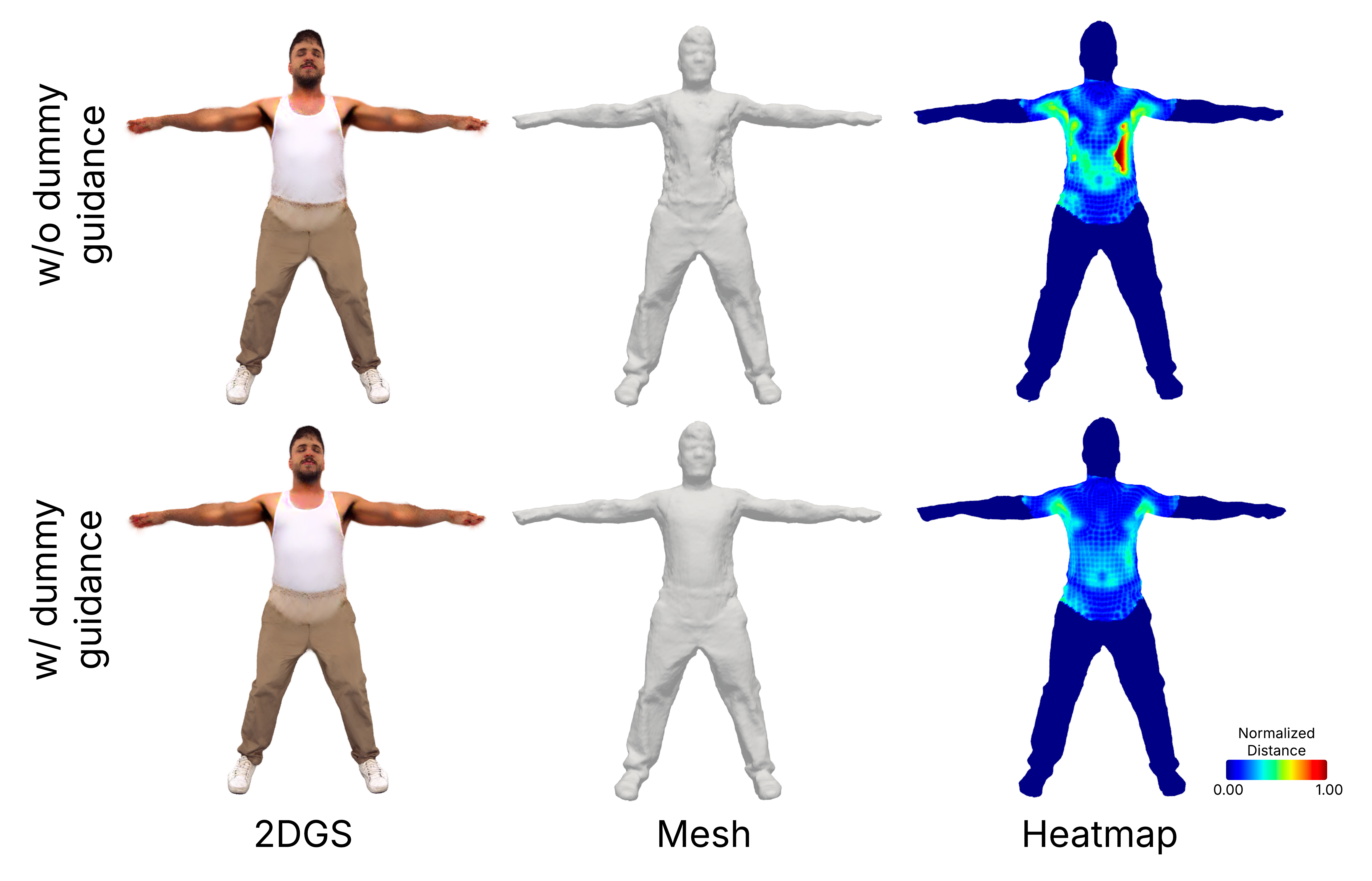}
        \caption{}
        \label{fig:ablation-dummy}
    \end{subfigure}
    \hfill
    \begin{subfigure}[b]{0.18\linewidth}
        \centering
        \includegraphics[width=\linewidth]{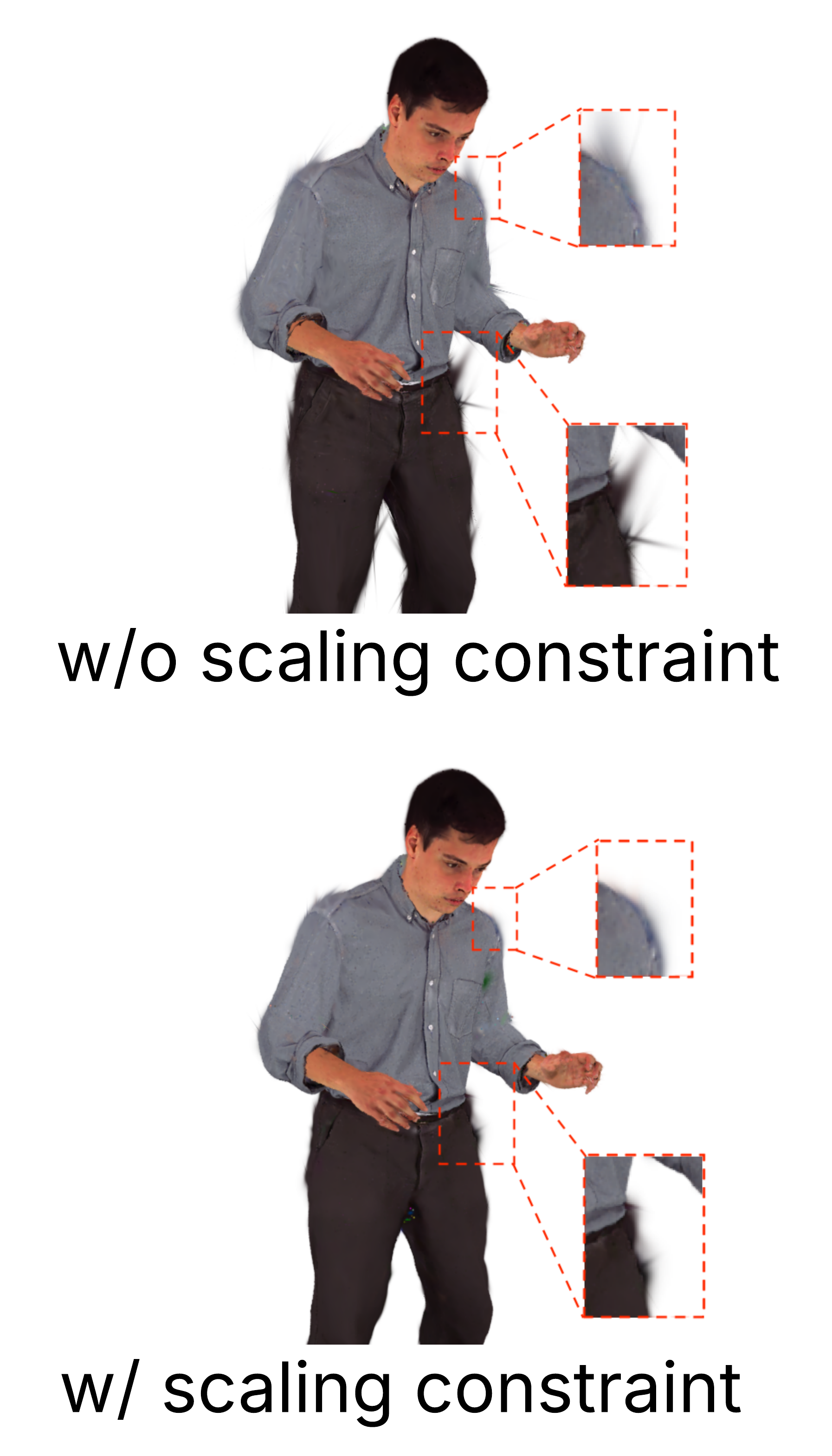}
        \caption{}
        \label{fig:ablation-scaling}
    \end{subfigure}
    \hfill
    \begin{subfigure}[b]{0.28\linewidth}
        \centering
        \includegraphics[width=\linewidth]{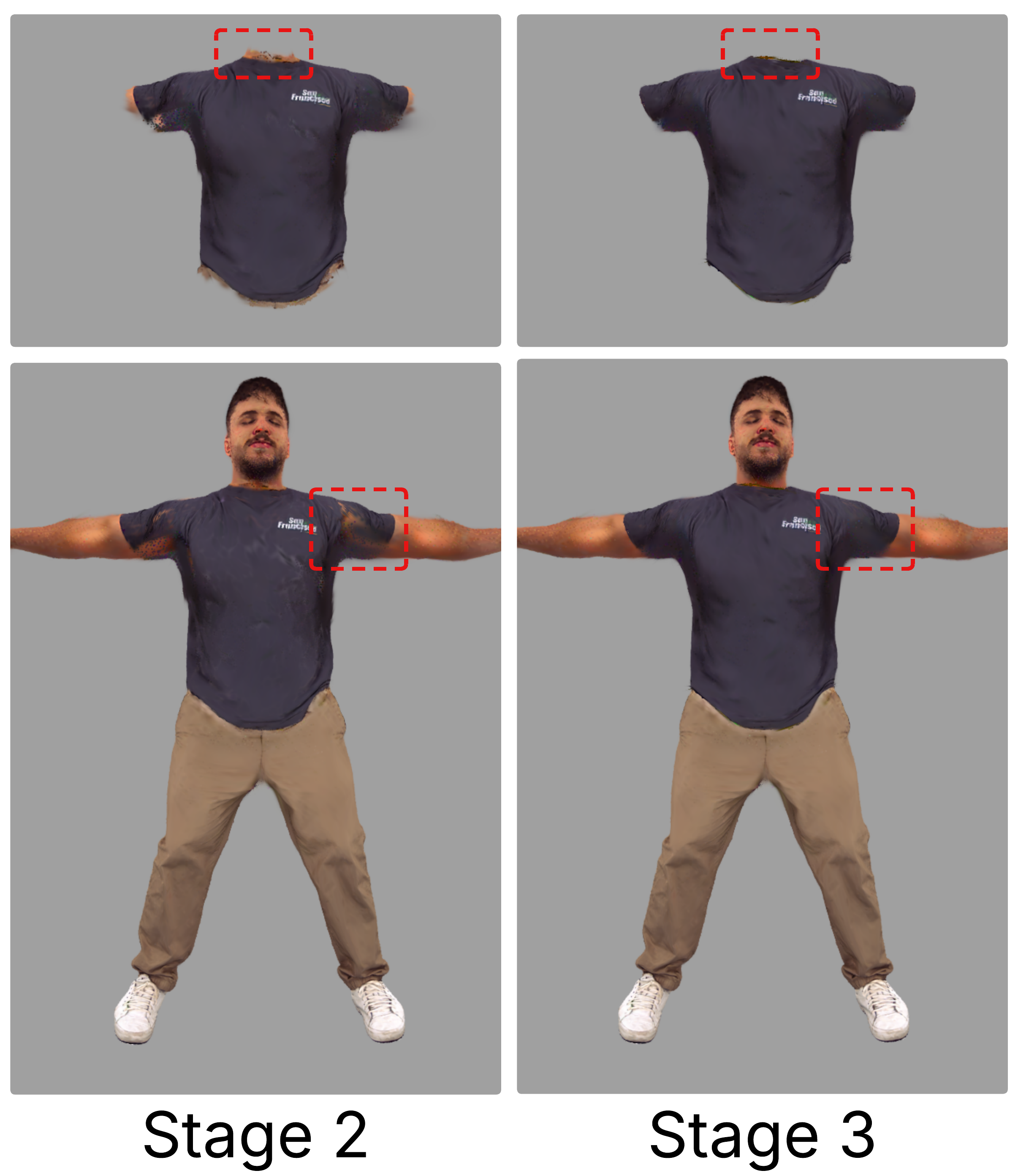}
        \caption{}
        \label{fig:ablation-stages}
    \end{subfigure}
    \captionsetup{font=footnotesize}
    \caption{\textbf{Ablations.} (a) Visualization of dummy shape guidance showing the rendered inner layer, mesh, and error heatmap. (b) Effect of the scaling constraint on virtual try-on results. (c) Comparison between Stage 2 and Stage 3.}
    \label{fig:full-ablation-study}
\end{figure*}

We conduct three ablation studies to evaluate the contributions of different components in our method. First, we compare Stage 2 and Stage 3. As illustrated in Fig.~\ref{fig:ablation-stages}, the refinement process removes inaccurately attached areas and densifies sparse outer-layer 2D Gaussians. Quantitative results in Table~\ref{tab:ablation-study-stage2-stage3} demonstrate improvements in CLIP and ImageReward scores, computed from a prompt describing the canonical composition of both layers. Second, we evaluate the effect of dummy SMPL(-X) body guidance. Fig.~\ref{fig:ablation-dummy} shows that, although the 2D Gaussian Splatting (2DGS) rendered outputs appear visually similar with or without dummy guidance, concave artifacts emerge in the mesh of the inpainted body areas when dummy guidance is not applied. A heatmap of mesh-to-ground-truth SMPL(-X) discrepancies shows reduced errors (Table~\ref{tab:ablation-study-guidance-difference}) with dummy guidance, which improves geometric accuracy and prevents rendering overlap in virtual try-on. Finally, we evaluate our scaling-constraint strategy. The original 2DGS training strategy, designed for static scenes, tends to generate large Gaussians for low-frequency areas. Since 2D Gaussians are attached to the mesh, when applying virtual try-on simulation under severe deformation (e.g., extreme stretching, sharp folding), large Gaussians lead to visual artifacts. Constraining Gaussian size during training effectively mitigates these artifacts (Fig.~\ref{fig:ablation-dummy}).

\begin{table}[t]
\centering
\footnotesize

\makebox[\linewidth][c]{%
  \begin{subtable}[t]{0.22\linewidth}
    \centering
    \begin{adjustbox}{max width=\linewidth}
      \begin{tabular}{l|c|c}
      \toprule
      Method & CLIP$\uparrow$ & IR$\uparrow$ \\
      \midrule
      Stage 2 & 30.83 & 1.378 \\
      Stage 3 & \textbf{30.96} & \textbf{1.388} \\
      \bottomrule
      \end{tabular}
    \end{adjustbox}
    \caption{}
    \label{tab:ablation-study-stage2-stage3}
  \end{subtable}
  \qquad

  \begin{subtable}[t]{0.35\linewidth}
    \centering
    \begin{adjustbox}{max width=\linewidth}
      \begin{tabular}{l|c|c|c}
      \toprule
      Method & Mean$\downarrow$ & Std$\downarrow$ & Max$\downarrow$ \\
      \midrule
      w/o guidance & 0.021 & 0.012 & 0.091 \\
      w/ guidance  & \textbf{0.015} & \textbf{0.006} & \textbf{0.048} \\
      \bottomrule
      \end{tabular}
    \end{adjustbox}
    \caption{}
    \label{tab:ablation-study-guidance-difference}
  \end{subtable}%
}
\captionsetup{font=footnotesize}
\caption{\textbf{Ablations.} (a) Quantitative results for garment refinement. (b) Heatmap differences with and without guidance, showing error metrics in occluded areas.}
\label{tab:ablation-studies-combined}
\end{table}
\section{Discussion}
We present a novel method for generating layered 3D human avatars from multi-view images of a statically posed subject. Leveraging 2D Gaussian Splatting guided by diffusion models, our approach enables the creation of photorealistic 3D garment and human assets, well-suited for applications such as virtual try-on. Experiments show that our method outperforms state-of-the-art decomposition work in both reconstructing and inpainting, while also being compatible with easily captured custom data, enhancing the practicality of the virtual try-on. 

\noindent\textbf{Limitations:} Our framework requires multi-view images of a static human. The decomposition may fail with loose garments or garments exhibiting self-intersections due to the dependence on SMPL(-X) linear blend skinning. The virtual try-on may also produce penetration artifacts when the garment is significantly smaller than the body. Extending the method to model and decompose a human from a single image is a promising direction for future work. Physics-based simulation could further improve overall robustness, and achieving high-fidelity rendering of gaussian garments under physical simulation and static poses remains an important area for future research.

\subsubsection{Acknowledgements} This work was conducted with the financial support of the Science Foundation Ireland Centre for Research Training in Digitally-Enhanced Reality (d-real) at Trinity College Dublin under Grant No. 18/CRT/6224. We also acknowledge the support from the Horizon Europe Framework Programme under Grant Agreement No. 101070109 (TRANSMIXR).

%
%
%
%

\end{document}